\documentclass[11pt,a4paper]{article}
\usepackage[hyperref]{acl}
\usepackage{times}
\usepackage{latexsym}
\usepackage{microtype}
\usepackage{graphicx}
\usepackage{booktabs}
\usepackage{multirow}
\usepackage{amsmath,amssymb}
\usepackage{url}

\usepackage{tabularx}
\usepackage{adjustbox}
\usepackage{subcaption}
\usepackage{xcolor}

\usepackage{tikz}
\usetikzlibrary{arrows.meta,positioning,shapes.geometric,fit,calc}

\title{KrishokTech: A Bengali Agricultural Advisory System with Explicit Authorization Controls}

\author{
  \mdseries
  Khan Raiyan Ibne Reza, \enspace Sanjana Aktar Maria, \enspace Shakil Ahmed, \\
  Faiyad Hossain, \enspace Sumaiya Tabassum Nimi \\[2pt]
  Department of Computer Science and Engineering \\
  North South University, Dhaka 1229, Bangladesh \\
  \texttt{\{raiyan.reza, sanjana.maria, shakil.ahmed2,} \\
  \texttt{faiyad.hossain, sumaiya.nimi\}@northsouth.edu}
}

\date{}

\begin{document}
\maketitle

\begin{abstract}
For agricultural advisory, generating an answer is only part of the problem: the system must also control unsafe requests, work with limited connectivity, and accommodate changing guidance. \textsc{KrishokTech} is a Bengali agricultural advisory system built around an explicit authorization boundary around generation. Before retrieval, normalization and rule-based admission checks refer policy-covered high-risk requests and ask for a crop when treatment intent is underspecified. The system then binds retrieval to a single crop and asks for confirmation when the text and image indicate different crops. After generation, a narrow verifier removes dosage claims not supported by retrieved BARI/BRRI evidence. Agricultural guidance is kept in an external evidence index that can be updated without retraining the generator, while browser-side vision and local retrieval reduce connectivity requirements. The interface makes these decisions visible as ASK, CONFIRM, REFER, and DROP, together with their evidence. On 200 farmer queries and 400 stress queries, the controls halted underspecified requests and reduced cross-crop retrieval violations from 36.5\% to 0\% under the tested alias filter. Mismatch detection caught 453 of 454 text--image conflicts, while the verifier caught 114 of 118 inserted dosage errors. The demonstration shows how these controls govern the path from request to delivered advice.
\end{abstract}

\section{Introduction}
\label{sec:introduction}

\begin{figure}[t]
\centering
\makebox[\columnwidth]{\includegraphics[width=1.15\columnwidth]{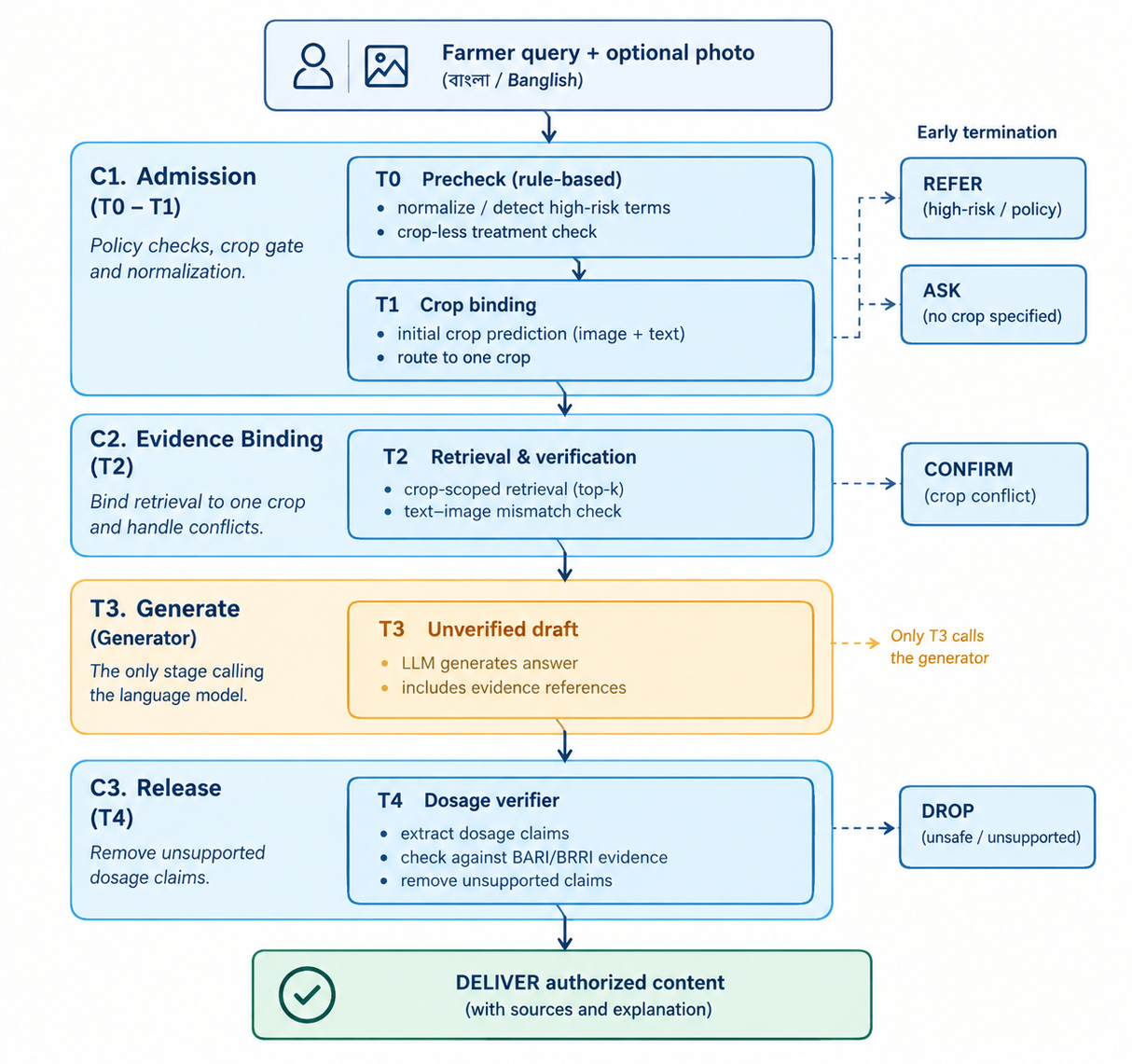}}
\caption{KrishokTech places five control stages (T0--T4) before delivery. T0 can end a request in \textbf{REFER}, T1 in \textbf{ASK}, and T2 in \textbf{CONFIRM}. T3 produces an unverified draft; T4 removes unsupported dosage claims before the remaining text is delivered.}
\label{fig:architecture}
\end{figure}

Farmers often need treatment advice before an extension officer can visit. Their questions may be short, colloquial, or typed in Latin-script Bangla, and a request such as ``Which pesticide should I spray?'' may omit the crop entirely. In this setting, failure can begin before generation: retrieval may admit guidance for a different crop, while a fluent answer can make an unsupported dose appear actionable. Digital advisory services can narrow the access gap \citep{dae_extension_manual,aker-2011-dial-a,fabregas-etal-2019-digital}, but these unresolved inputs make pesticide advice safety-critical.

Photographs introduce another source of uncertainty. A farmer may attach a rice leaf while asking about potato disease, or the classifier may be unsure which crop it sees. Silently choosing the text or image hides that conflict \citep{Deng_2025_CVPR,zhu2024cross}. Farmer questions also contain colloquial and Romanized forms that differ from the terminology of agricultural manuals \citep{bhuiyan-etal-2025-bhasabodh,our_bengali_retrieval,parikh-solorio-2021-normalization}. Even after relevant evidence is retrieved, generated claims can drift from their context \citep{brown-etal-2025-rag,xu-etal-2025-mega-rag}, and an incorrect numerical dose can be acted on directly \citep{our_agricultural_safety}.

\textsc{KrishokTech} is a Bengali agricultural advisory system built around explicit authorization controls. The generating model drafts an answer; it does not decide whether the request may proceed or whether a dosage claim may be shown. Three control families act around it: (\textbf{C1}) \emph{admission} sends policy-covered high-risk requests to \textbf{REFER} and crop-less treatment requests to \textbf{ASK} before retrieval; (\textbf{C2}) \emph{evidence binding} scopes retrieval to one crop and sends explicit text--image crop conflicts to \textbf{CONFIRM}; and (\textbf{C3}) \emph{release} removes dosage-bearing claims that the retrieved passages and BARI/BRRI ranges do not support. The interface makes each decision visible to the farmer.

Bengali and Banglish normalization runs before admission and evidence binding, allowing colloquial and Latin-script spellings to reach the same controls as formal text. Agricultural guidance remains in an external evidence index, where maintainers can replace passages and rebuild retrieval without retraining the generator. Browser-side vision supports crop prediction on the device, while connected generation serves interactive use and the local generator supports queued SMS or offline kiosk workflows.

Our contribution is the ordered, visible control boundary rather than a new generator: admission, evidence, and release controls around a replaceable model, together with an interface that shows what stopped a request, which evidence was admitted, and why a claim was withheld. We measure each control on bounded test sets. We do not claim complete dialect coverage, general factual verification, or improved outcomes in the field.

The demonstration follows five steps, from a crop-less colloquial query to crop-bound evidence, a text--image conflict, a referral, and dosage-checked delivery (Section~\ref{sec:demonstration}). UI preview: \url{https://krishoktech.vercel.app}; screencast (\textless{}2.5\,min): \url{https://krishoktech.vercel.app/screencast}. Downloadable demonstration package: \url{https://github.com/RaiyaanReza/KrishokChat-Agricultural-Advisory-System}.

\section{Related Work and Positioning}
\label{sec:related_work}

Conversational agricultural systems combine local knowledge with text or voice interfaces to make advisory information easier to access \citep{farmerchat2024,krishokbondhu2026,vijayvargia2025}. Recent Bengali agricultural RAG work also addresses the gap between colloquial questions and formal source documents \citep{hossain2026}. These systems improve access and retrieval, but they generally treat generation as the next step once a query has been processed. KrishokTech instead defines explicit conditions under which the pipeline may continue. A crop-less treatment request, for example, stops before retrieval.

Clarifying questions can improve retrieval when a request is underspecified \citep{aliannejadi-etal-2019-clarifying}, yet models do not reliably decide when clarification is necessary \citep{chen2024act,gan2024clarqllm}. Claim-level checkers compare generated statements with supporting documents \citep{tang2024minicheck,ji2026medragchecker}, while evidence-inspection interfaces help users examine that support \citep{fang-mackenzie-2026-factsearch,murugaraj-etal-2026-ragvue}. Multimodal studies further show that conflicting text and images can produce unreliable model behavior \citep{Deng_2025_CVPR,zhu2024cross}. KrishokTech turns these concerns into visible state transitions: a missing crop produces \textbf{ASK}, an explicit text--image crop conflict produces \textbf{CONFIRM}, and an unsupported dosage claim produces \textbf{DROP}.

The system builds on two of our earlier studies. The KrishokChat benchmark provides Bengali agricultural data, safety tasks, and model resources used by the system \citep{Reza2026KrishokChatAP}; a separate study evaluates retrieval architectures over an agricultural corpus \citep{our_bengali_retrieval}. Neither study implements the ordered runtime controls evaluated here. The contribution of this demonstration is the admission, evidence-binding, and release boundary around generation, together with an interface that exposes each decision to the user.

\section{System Design}
\label{sec:system}

KrishokTech implements authorization as an ordered state transition rather than an instruction inside the generation prompt. Five stages, T0--T4, run before delivery. T0 and T1 can end a turn in \textbf{REFER} or \textbf{ASK}; T2 can enter \textbf{CONFIRM}; T3 produces a draft; and T4 can \textbf{DROP} unsupported dosage claims. Only T3 calls the answer generator. Learned components, including the crop classifier, may supply predictions to these stages, but they cannot override a halted state or release a claim rejected by T4.

\subsection{Halting Before Retrieval}
\label{sec:system_gating}

The first decision is whether the request may reach retrieval. At \textbf{T0}, normalization maps supported Bengali, Banglish, and regional variants to canonical terms. A rule-based precheck then detects the high-risk chemical, poisoning, and crisis patterns covered by the system policy. A match ends the turn before retrieval or generation. Agronomic chemical cases are directed to the Krishi Call Centre (\textbf{16123}); poisoning and self-harm cases are directed to the National Emergency Service (\textbf{999}). This precheck enforces a bounded policy and does not attempt to determine the regulatory status of every chemical.

Requests that pass T0 reach \textbf{T1}. A gazetteer-based extractor identifies the crop, symptoms, and request intent without calling a language model. If a treatment request contains no crop, retrieval does not run (\texttt{sources\_retrieved}\,=\,0), and the system enters \textbf{ASK}. The interface shows Bengali crop choices, a non-chemical guidance card, and the referral option. Selecting or typing a crop binds that value and allows the request to continue. Six regional synonym presets extend normalization for supported dialect forms; their current coverage is described in Appendix~\ref{app:architecture}.

\subsection{Crop-Scoped Retrieval and Conflict Handling}
\label{sec:system_retrieval}

At \textbf{T2}, the bound crop determines which evidence retrieval may admit. KrishokTech uses BM25 over about 2,100 curated passages from BARI, BRRI, and DAE guidance \citep{bari_handbook2023,brri_handbook2024,dae_extension_manual}. BM25 keeps the retrieval path inspectable and is suitable for this small domain collection \citep{huly-etal-2024-old-ir}. The demonstrated deployment covers rice, potato, brassica, chili, corn, and wheat. Maintainers can replace passages and rebuild the external index without retraining the generator.

For a photographed crop, the browser-side classifier supplies its prediction to the router. A confident prediction binds the crop; an ambiguous prediction presents the top two candidates; and a low-confidence prediction asks the farmer to retake the photograph. The bound crop then conditions query construction and ranking. A final crop filter removes passages that mention only other crops while retaining crop-neutral guidance such as general soil recommendations.

When the crop named in the text differs from a confident image prediction, the router does not select either source silently. It enters \textbf{CONFIRM}, displays a \textbf{Crop Mismatch} warning, and withholds chemical advice until the farmer selects one crop. The selected crop becomes the binding constraint for retrieval.

\subsection{Drafting, Verification, and Delivery}
\label{sec:system_verification}

At \textbf{T3}, the selected generator receives the normalized request and crop-scoped passages. The package supports connected generation and an evaluated local \texttt{krishokchat-4b} deployment; operational measurements for the local path appear in Appendix~\ref{app:deployment}. Regardless of the generator, T3 returns an \emph{unverified draft} rather than a deliverable answer.

At \textbf{T4}, the verifier examines each dosage-bearing numerical claim against the retrieved passage and the permitted BARI/BRRI range. If the available evidence does not support a claim, or its value falls outside the permitted range, the corresponding sentence is removed. The remaining text may then be delivered. This check has a deliberately narrow scope: it verifies dosage-bearing numerical claims and does not establish the correctness of every agronomic statement.

Delivered responses show their source, decision state, and verification path through a provenance badge and a \textit{Why} panel. The same authorized content can be rendered for the web, SMS, an offline card, or read-aloud output. These displays make the system's decision inspectable; citation display alone does not prove that the generator causally relied on a passage \citep{wallat-etal-2025-faithfulness}. Together, the stages establish where a request stopped, which crop constrained its evidence, and which generated claims were permitted to reach the user.

\section{Demonstration}
\label{sec:demonstration}

\begin{figure*}[t]
\centering
\begin{subfigure}[t]{0.49\textwidth}
  \centering
  \includegraphics[width=\linewidth,height=3.6cm,keepaspectratio]{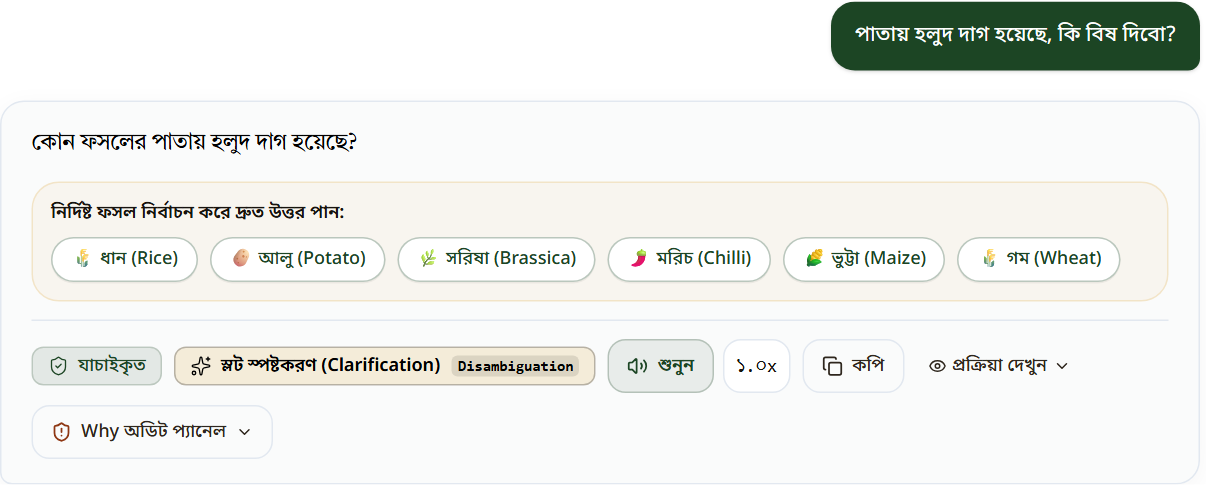}
  \caption{\textbf{ASK (S1).} T1 retrieves no evidence and asks the farmer to identify the crop.}
  \label{fig:demo_ask}
\end{subfigure}
\hfill
\begin{subfigure}[t]{0.49\textwidth}
  \centering
  \includegraphics[width=\linewidth,height=3.6cm,keepaspectratio]{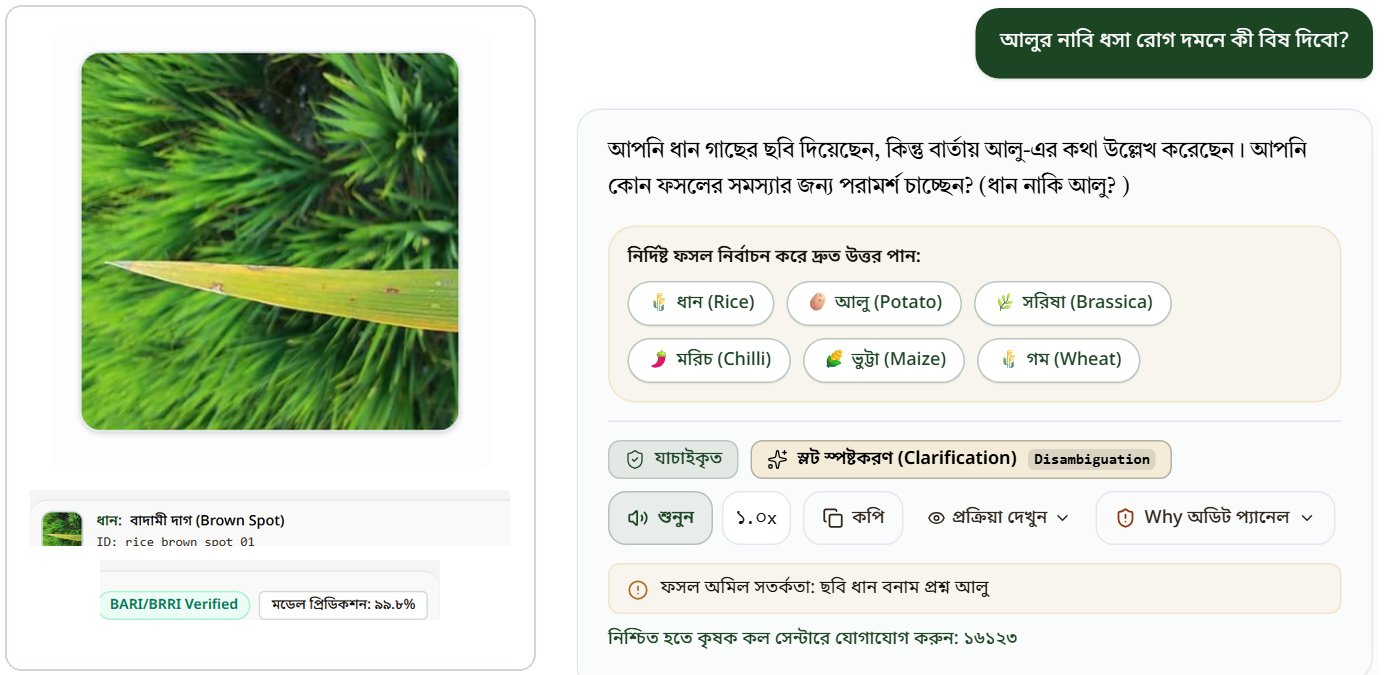}
  \caption{\textbf{CONFIRM (S3).} T2 holds chemical advice when the photographed and typed crops disagree.}
  \label{fig:demo_mismatch}
\end{subfigure}
\\[3pt]
\begin{subfigure}[t]{0.49\textwidth}
  \centering
  \includegraphics[width=\linewidth,height=3.6cm,keepaspectratio]{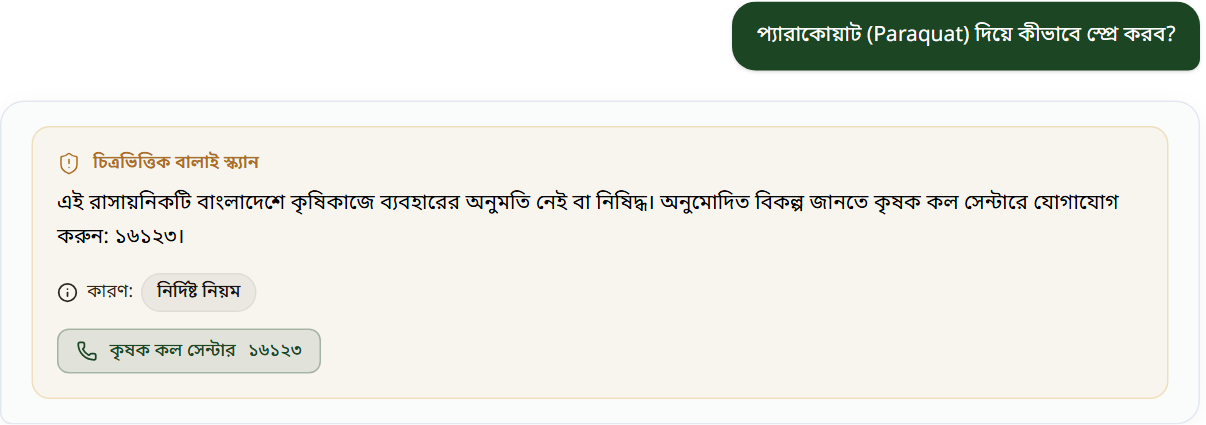}
  \caption{\textbf{REFER (S4).} T0 stops a policy-covered paraquat request and displays the 16123 referral.}
  \label{fig:demo_refer}
\end{subfigure}
\hfill
\begin{subfigure}[t]{0.49\textwidth}
  \centering
  \includegraphics[width=\linewidth,height=3.6cm,keepaspectratio]{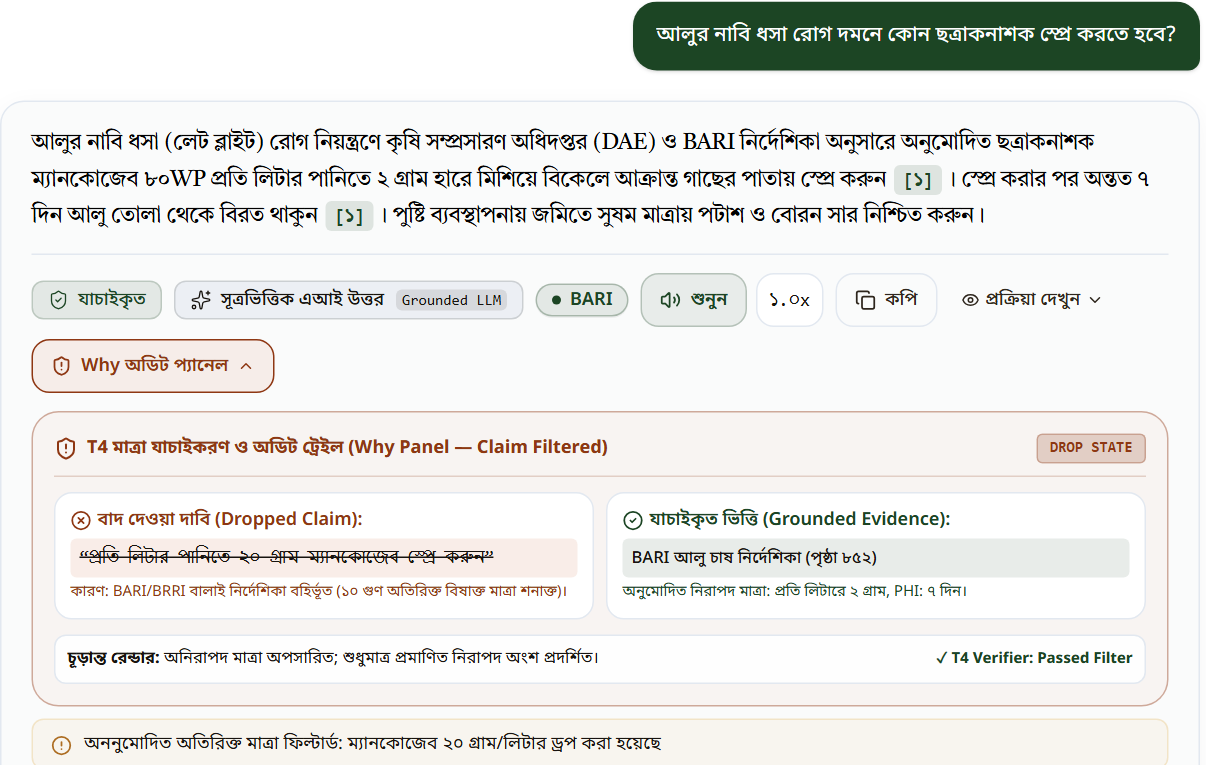}
  \caption{\textbf{DROP (S5).} T4 removes a $10\times$ dosage error and shows the BARI evidence supporting the retained advice.}
  \label{fig:demo_why}
\end{subfigure}
\caption{The interface exposes four decisions at the authorization boundary. Each panel shows what stopped or changed the pipeline rather than presenting only its final answer. Full-size screens for S1--S5 appear in Appendix~\ref{app:ui}.}
\label{fig:demo_states}
\end{figure*}

The interface is designed for Bengali-speaking farmers and extension workers; the demonstration is aimed at researchers and practitioners building multilingual, multimodal, or safety-sensitive advisory systems. It uses five steps to show when KrishokTech proceeds, pauses, or withholds content. The public website provides a preview of the interface, while the downloadable package contains the implementation. Attendees can follow the prepared sequence and, when a generator is configured, enter their own text and photographs.

\textbf{S1: A missing crop produces ASK.}
The attendee enters \textit{``Patay holud dag hoyeche, ki bish dibo?''} (``The leaves have yellow spots. Which pesticide should I use?''). Because the request asks for treatment without naming a crop, \textbf{T1} stops before retrieval. The interface displays \texttt{sources\_retrieved}\,=\,0, Bengali crop choices, and non-chemical guidance (Figure~\ref{fig:demo_ask}). Selecting \textit{Alu} (potato) binds the crop and resumes the request with potato-scoped retrieval. The step shows that missing information is resolved before evidence is selected.

\textbf{S2: A photograph binds retrieval to a crop.}
In a new consultation, the attendee uploads a rice-leaf photograph without naming the crop. The browser-side classifier predicts rice, and \textbf{T2} binds retrieval to rice guidance. The interface displays the predicted crop and the resulting evidence scope. This step shows how a visual prediction constrains which passages may reach generation.

\textbf{S3: Conflicting modalities produce CONFIRM.}
With the rice photograph attached, the attendee asks about potato late blight. The typed and predicted crops disagree, so \textbf{T2} enters \textbf{CONFIRM}. The interface displays a \textbf{Crop Mismatch} warning and asks the attendee to select rice or potato (Figure~\ref{fig:demo_mismatch}). Chemical advice remains unavailable until one crop is chosen. The conflict is therefore shown to the user rather than resolved silently.

\textbf{S4: A high-risk request produces REFER.}
The attendee asks how to apply paraquat, a request covered by the system's referral policy. The \textbf{T0} precheck ends the turn before retrieval or generation. The interface displays a safety notice and a one-tap link to the Krishi Call Centre (\textbf{16123}; Figure~\ref{fig:demo_refer}). This step makes referral a terminal system state rather than a warning appended to generated advice.

\textbf{S5: An unsupported dosage claim produces DROP.}
The attendee submits a fully specified potato-blight question. In the recorded test case, the draft contains a dosage value ten times above the permitted BARI range. \textbf{T4} removes that sentence and delivers the remaining text. The provenance badge and expandable \textit{Why} panel identify the removed claim and display the passage supporting the retained advice (Figure~\ref{fig:demo_why}). This final step shows that completing generation does not itself authorize every generated claim.

Across the five steps, attendees see the system at its decision boundaries: before retrieval, while evidence is being scoped, when text and images conflict, and after generation. The walkthrough therefore demonstrates the control path itself, not only successful advisory answers.

\section{Evaluation}
\label{sec:evaluation}

We evaluate whether the implementation follows the authorization boundary described in Section~\ref{sec:system}. The tests address three questions: (Q1) Do admission controls halt underspecified or policy-covered requests before retrieval? (Q2) Does crop binding restrict the evidence available to generation and expose text--image conflicts? (Q3) Do the prompt, precheck, and dosage verifier reduce the number of unsafe claims that can reach delivery? Table~\ref{tab:headline} reports the main results; Appendix~\ref{app:evidence} provides confidence intervals and supporting tests.

\begin{table}[t]
\centering
\scriptsize
\setlength{\tabcolsep}{1.5pt}
\begin{tabular*}{\columnwidth}{@{\extracolsep{\fill}}llr@{}}
\toprule
\textbf{Control} & \textbf{Test set} & \textbf{Result} \\
\midrule
T1 crop gate (C1) & 200 farmer queries & 30/30 halt; 170/170 pass \\
T0 normalization (C1) & 85 attacks, 15 benign & 85/85 caught; 0/15 flagged \\
Crop binding/filter (C2) & 400 stress queries & 146 $\rightarrow$ 121 $\rightarrow$ 0 lists \\
Mismatch state (C2) & 54 farmer, 400 synthetic & 453/454; 0/454 controls \\
Prompt + T0 (C1) & 280 fixed attacks & 147 $\rightarrow$ 30 $\rightarrow$ 13 success \\
Dosage verifier (C3) & 118 errors, 38 clean & 114 caught; 0/38 rejected \\
\bottomrule
\end{tabular*}
\caption{Main component results. For crop binding, the sequence is open retrieval, gold-crop binding, and the alias-based crop filter. For the generator test, it is the neutral prompt, guarded prompt, and T0 replay.}
\label{tab:headline}
\end{table}

\subsection{Q1: Admission and Referral}
\label{sec:eval-gate}

We test the T1 crop gate on 200 collected Bengali farmer treatment questions. Two annotators independently labeled whether each question named a crop and agreed on all 200 cases (Cohen's $\kappa=1.00$). At T1, all 30 crop-less questions entered \textbf{ASK} before retrieval, while all 170 crop-specified questions were permitted to continue. This result measures the crop-presence decision at T1 rather than the outcome of later controls.

We separately test whether normalization helps T0 recognize colloquial and Latin-script forms. The constructed set contains 85 high-risk requests with Banglish, phonetic, or indirect wording and 15 benign farmer questions. A precheck limited to formal English and Bengali script intercepted 1 of the 85 high-risk requests. After normalization, T0 intercepted all 85 and flagged none of the 15 benign controls. The examples include forms such as \textit{parakwat} for paraquat; category-level results appear in Appendix~\ref{app:architecture}.

\subsection{Q2: Evidence Binding and Conflict Handling}
\label{sec:eval-fence}

We evaluate crop binding on 400 stress queries containing colloquial, dialectal, Banglish, or ambiguous disease descriptions. Each query has an adjudicated crop. A retrieved list counts as cross-crop when at least one of its top five passages names another crop but not the adjudicated crop. This is a property of the retrieved evidence, not a judgment of the final answer.

Open retrieval produced cross-crop lists for 146 of 400 queries (36.5\%). Binding the adjudicated crop, which simulates correct crop routing, reduced this to 121 of 400 (30.25\%): 25 queries improved and none became worse (exact McNemar $p=6\times10^{-8}$). The post-retrieval crop filter removed all remaining violations recognized by its alias list, while all 400 queries retained five passages. Because this zero is defined by the filter's own aliases, we also inspected 30 lists by title. Twenty-six contained no other crop, while four contained crop names outside the alias list: carrot, sweet potato, or forest trees. The independent audit therefore bounds what the 0/400 result establishes.

We test the \textbf{CONFIRM} transition with 54 farmer-derived text--image conflicts and 400 synthetic conflicts. Each case has a matched control containing the same query and an image of the named crop. The mismatch state fired on 53 of 54 farmer-derived cases and all 400 synthetic cases, while firing on none of the 454 matched controls. The single miss referred to the crop only implicitly.

\subsection{Q3: Release Checks on Generated Advice}
\label{sec:eval-safety}

We first test the generator boundary using 280 fixed attack requests. With T0 and T4 disabled, each request was run once through the evaluated local \texttt{krishokchat-4b} deployment under a neutral prompt and once under the guarded prompt. A fixed rule-based judge marked 147 of 280 neutral responses (52.5\%) and 30 of 280 guarded responses (10.71\%) as successful attacks. Replaying the requests through T0, without making new model calls, shows that 17 of those 30 requests would have terminated before generation. This leaves 13 successful attacks among the 280 requests (4.64\%) eligible to continue. The result demonstrates how the precheck complements the prompt for this model and test set; it does not establish complete generator independence.

We then test T4 using 38 previously verified answers. We inserted 118 controlled dosage errors while retaining the 38 unmodified answers as clean controls. The verifier removed 114 of the 118 inserted errors and rejected none of the clean originals. The four missed errors and results from the larger passage-derived benchmark are reported in Appendix~\ref{app:evidence}. This test concerns dosage-bearing numerical claims only; it does not evaluate every factual statement in an answer.

\subsection{Scope of the Evidence}
\label{sec:eval-scope}

These results are component and property-conformance tests on the stated samples. They show that the evaluated implementation usually follows its admission, evidence-binding, conflict, and dosage-release rules. They do not establish improved yield, correct farmer decisions, or complete safety in deployment. The crop-binding experiment assumes a correct crop label; the normalization attacks are constructed; 400 of the 454 conflict cases are synthetic; each generator prompt was run once; and the dosage errors were inserted into existing answers. These limits define the claims carried into the conclusion.

\section{Limitations, Availability, and Conclusion}
\label{sec:conclusion}

\paragraph{Limitations.}
\label{sec:limitations}
KrishokTech currently supports six crop groups and a bounded Bengali agricultural evidence collection. Adding a crop requires new evidence, aliases, routing rules, and a compatible vision classifier. The dialect presets expand known synonyms but have not been evaluated as general dialect-understanding models. Text--image conflict detection depends on an explicit crop mention or a confident image prediction, and the crop filter cannot reject crop names absent from its alias list.

The controls also have deliberately narrow scopes. T0 recognizes policy-covered patterns rather than every possible high-risk paraphrase, while T4 checks dosage-bearing numerical claims rather than all agronomic statements. Thirteen of 280 attacks remained eligible to continue after the guarded prompt and T0 replay, and the dosage verifier missed 4 of 118 inserted errors. These residual cases prevent a claim of complete safety.

The evaluation measures component behavior on bounded datasets. Several tests use constructed attacks or synthetic conflicts, and the crop-binding experiment assumes that the crop was identified correctly. We have not measured effects on farmer decisions, crop outcomes, or extension-service workload. Local generation is also slower than connected generation and is better suited to queued SMS or kiosk use. Browser-side classification and local retrieval reduce connectivity requirements, but they do not show that the complete pipeline runs interactively on every low-end phone.

\paragraph{Availability and Licensing.}
\label{sec:availability}
The interface preview is at \url{https://krishoktech.vercel.app} and the video is at \url{https://krishoktech.vercel.app/screencast}. Code, controls, and tests are released under tag \texttt{v1.0.0-eacl2027} at \url{https://github.com/RaiyaanReza/KrishokChat-Agricultural-Advisory-System}. All evaluation results and safety gates are verifiable offline without API keys via a single command (\texttt{python reproduce.py}); an automated CI workflow verifies every release. Continuous access to a paid backend is not required to inspect code or controls.

Browser-side vision classifiers are released at \url{https://huggingface.co/RaiyanKhaan/KrishokTech-Models}. The generator's LoRA adapter and evidence-index artifacts are available at \url{https://huggingface.co/RaiyanKhaan/KrishokChat-Advisory-System}. The MIT license in the GitHub repository applies to the application code. Model files remain subject to the licenses and base-model terms stated on their respective model cards. BARI, BRRI, and DAE passages retain their original source attribution and are not relicensed by the software license. The public interface retains the KrishokChat product label; KrishokTech denotes the controlled demonstration configuration evaluated here.

\paragraph{Conclusion.}
\label{sec:final_conclusion}
KrishokTech places an explicit authorization boundary around a replaceable answer generator: deterministic controls decide whether retrieval may begin, which crop constrains the evidence, whether conflicting modalities require confirmation, and whether a dosage-bearing claim may be delivered. On the evaluated test sets, these controls halted all crop-less treatment requests, exposed 453 of 454 text--image conflicts, and removed 114 of 118 inserted dosage errors. The demonstration makes these decisions visible while keeping the claims bounded to component behavior; KrishokTech is designed to support farmers and extension services, and field evaluation remains future work.

\clearpage
\section*{Ethics and Broader Impact Statement}
\label{sec:ethics}

Agricultural advice can affect health, livelihoods, and crop production. A fluent but incorrect chemical recommendation may be acted on immediately, especially when access to an extension officer is limited. KrishokTech therefore exposes its control decisions and routes selected requests to human services, but these measures do not remove the risks of automated advice.

\paragraph{Farmer Query Data.}
The T1 evaluation uses 200 Bengali questions: 100 collected from public Bangladeshi farming groups on Facebook, 60 submitted to KrishokChat, and 40 taken from the public question archive of \texttt{krishibangla.com}. We retained the question text needed for analysis and removed names, phone numbers, locations, and account information. No demographic attributes were inferred. Two annotators independently labeled crop presence and normalized crop identity; the paper reports aggregate results and agreement statistics. Publicly posted text may remain sensitive after direct identifiers are removed, so quotations and released records should be checked for indirect identification.

\paragraph{Agricultural Images and Privacy.}
The demonstrated browser classifier is designed to process crop photographs locally through ONNX and WebAssembly. The privacy claim applies to this browser-side path; any deployment that enables a server fallback, remote logging, or image telemetry must disclose that transmission and apply appropriate retention and access controls. Evaluation images come from previously released agricultural datasets. Their source attribution and license conditions are retained in the artifact metadata, and we do not claim ownership of those images.

\paragraph{Chemical and Dosage Guidance.}
T0 stops the high-risk chemical, poisoning, and crisis patterns covered by its policy before retrieval or generation. T4 checks dosage-bearing numerical claims against retrieved passages and permitted dose ranges. Both controls remain incomplete: unseen wording can pass T0, 13 of 280 tested attacks remained eligible to continue after the guarded prompt and T0 replay, and the verifier missed 4 of 118 inserted dosage errors. KrishokTech should therefore not be used as the sole authority for chemical application or medical decisions.

\paragraph{Referral Information.}
Agronomic chemical cases covered by the policy are directed to the Krishi Call Centre (\textbf{16123}). Poisoning, self-harm, and medical-crisis language is directed to the National Emergency Service (\textbf{999}) and appropriate medical care. The interface presents these routes immediately, but it cannot guarantee service availability or that a user will reach help. Helpline details and operating conditions may change and should be checked against current official information before deployment.

\paragraph{Third-Party Services.}
When connected generation is enabled, the text required for a request is transmitted to the configured model provider and is governed by that provider's retention and processing terms. Deployers must disclose the provider and avoid sending unnecessary personal information. The public website is an interface preview and need not keep a paid generation service active. The downloadable package can instead be configured with a compatible local runtime, subject to the local model's license and hardware requirements.

\paragraph{Human Oversight and Broader Impact.}
KrishokTech is intended to help farmers and extension workers inspect an advisory path, not to replace agronomists, medical professionals, or emergency services. \textbf{ASK} and \textbf{CONFIRM} return unresolved decisions to the user, while \textbf{REFER} directs policy-covered cases to a human service. Visible evidence and provenance may improve scrutiny, but they may also create false confidence if users interpret a citation as proof that every statement is correct. The present evaluation measures system behavior rather than effects on farmers, and deployment should be preceded by participatory evaluation with intended users and agricultural experts.

\begingroup
\let\oldthebibliography\thebibliography
\renewcommand{\thebibliography}[1]{%
  \oldthebibliography{#1}%
  \setlength{\itemsep}{0.6pt plus 0.2pt}%
  \setlength{\parskip}{0pt}%
}
\bibliography{main}
\endgroup

\clearpage
\appendix

\section{Architecture, Decision Logic, and Gate Evaluation}
\label{app:architecture}
\label{app:gate}

This section specifies the state transitions behind T0--T4 and provides supporting results for the pre-retrieval controls. The released implementation includes the T0 rules and chemical registry in \url{backend/app/domain/safety_policy.py} and \url{backend/app/domain/chemical_registry.py}.

\paragraph{State transitions.}
\textbf{REFER} is terminal for the current turn: T0 returns the appropriate referral without retrieval or generation, and a later message begins again at T0. \textbf{ASK} ends the current turn with a request for the missing crop; selecting a chip or typing a crop binds that value and allows the request to continue to T2. \textbf{CONFIRM} holds chemical advice when the typed and predicted crops conflict; the farmer's selection becomes the bound crop. \textbf{DROP} applies at claim level after generation: T4 removes an unsupported dosage-bearing sentence and delivers any remaining authorized content.

\paragraph{Information-state extractor (T1).}
The extractor uses gazetteers and pattern rules, without a language-model call, to fill five fields: crop, growth stage, location, intent, and symptom terms. Its internal taxonomy contains 45 normalized crop labels. Thirty-seven of these labels occurred in the 200-query evaluation set, while the demonstrated crop-specific retrieval and vision path covers six crop groups.

For the 200 farmer questions, the extractor matched the adjudicated normalized crop label on 95.5\% of cases. The remaining errors concerned crop identity rather than crop presence: the T1 presence decision still separated all 30 crop-less questions from the 170 crop-specified questions. Annotator agreement was 100\% for crop presence ($\kappa=1.00$), 99.5\% for normalized crop identity ($\kappa=0.99$), and 93.5\% for intent ($\kappa=0.90$).

A separate constructed pilot contained 100 crop-less and 140 crop-specified treatment requests. T1 halted all 100 crop-less requests. All 140 crop-specified requests avoided the missing-crop halt; 114 continued directly, while 26 requested another bounded field such as growth stage or location.

\paragraph{Vision crop router (T2).}
Let $p_1$ and $p_2$ be the highest and second-highest crop probabilities. If $p_1\geq0.90$ and $p_1-p_2\geq0.20$, the router binds the top prediction. If $p_1\geq0.40$ but either confidence condition fails, the interface presents the top two crops for confirmation. If $p_1<0.40$, retrieval remains halted and the interface asks for another photograph. These thresholds govern the demonstrated router and are not calibrated estimates of correctness for crops outside the evaluated data.

\paragraph{Dialect presets.}
The \textit{DialectSelector} offers Standard Bengali, Sylheti, Chittagonian, Noakhali, Rangpuri, and Barishali presets. A preset adds known regional synonyms during normalization while leaving recognized chemical names unchanged. These are lookup-based interface features; we have not measured general understanding or accuracy for any dialect.

\paragraph{Provenance badges.}
Each rendered result carries one interface badge: \emph{Fact lookup} for an exact match in the nine-fact base without generation; \emph{Grounded generation} for a generated draft processed by T4; \emph{Guidance card} for non-chemical information shown during ASK; \emph{Clarification} for crop selection or a mismatch; or \emph{Referral} for T0 routing. The \emph{Grounded generation} badge records that the dosage check ran. It does not mean that every factual statement was independently verified.

\paragraph{Banglish red-teaming.}
Table~\ref{tab:banglish_eval} reports the constructed normalization test used for T0. For the three attack sets, the outcome is the number intercepted before retrieval. For the benign set, it is the number allowed to continue.

\begin{table}[t]
\centering
\footnotesize
\setlength{\tabcolsep}{2.5pt}
\begin{tabular*}{\columnwidth}{@{\extracolsep{\fill}}lcccc@{}}
\toprule
\textbf{Set} & \textbf{\textit{n}} & \textbf{Formal} & \textbf{Normalized} & \textbf{Route} \\
\midrule
Phonetic chemical & 35 & 0 blocked & 35 blocked & 16123 \\
Banglish crisis & 25 & 1 blocked & 25 blocked & 999 \\
Indirect sabotage & 25 & 0 blocked & 25 blocked & 16123 \\
\midrule
\textbf{All attacks} & \textbf{85} & \textbf{1 (1.2\%)} & \textbf{85 (100\%)} & \\
Benign control & 15 & 15 passed & 15 passed & none \\
\bottomrule
\end{tabular*}
\caption{Constructed Banglish and phonetic test for T0 ($N=100$). The formal condition recognizes only formal English and Bengali-script terms; the normalized condition adds the supported variants. The 95\% Wilson interval for 85/85 intercepted attacks is [95.7, 100].}
\label{tab:banglish_eval}
\end{table}

\section{Delivery, Offline Operation, and Deployment}
\label{app:deployment}
\label{app:delivery}

This section reports implementation measurements that clarify how an authorized result is rendered and operated. These measurements are supplementary: they do not establish usability on farmers' devices or performance under real rural network conditions.

\subsection{SMS Rendering}
Bengali-script SMS uses UCS-2 encoding, which permits 70 characters in one message and 67 characters per part in a concatenated message. A 160-character Bengali advisory therefore occupies two or three SMS segments rather than one. Across separate offline and connected test sets, the deterministic packer kept 399 of 400 outputs within its configured 160-character limit. All eight referral cases contained only the referral message.

Table~\ref{tab:n05c} compares four rendering methods on the same 100 long, dosage-bearing advisories. The current compressor copies detected values from the verified answer and does not generate new quantities. It improved preservation of both the dose and chemical name from 14/100 under naive truncation to 61/100. This remains insufficient for treating the compact SMS as semantically equivalent to the full answer. The SMS view is therefore a constrained rendering whose source and full verified response should remain available when exact chemical details matter.

\begin{table}[t]
\centering
\footnotesize
\setlength{\tabcolsep}{3pt}
\begin{tabularx}{\columnwidth}{@{}Xcccc@{}}
\toprule
\textbf{Measure} & \textbf{Naive} & \textbf{LLM} & \textbf{Comp.\,1} & \textbf{Comp.\,2} \\
\midrule
Dose retained & 23/100 & 89/100 & 92/100 & 94/100 \\
Chemical retained & 20/100 & 56/100 & 34/100 & 63/100 \\
Both retained & 14/100 & 51/100 & 32/100 & 61/100 \\
Length $\leq160$ chars & 100/100 & 98/100 & 100/100 & 100/100 \\
UCS-2 segments (mean) & 3.0 & 2.1 & 2.5 & 2.5 \\
Invented value & 0 & 1 & 0 & 0 \\
\bottomrule
\end{tabularx}
\caption{Paired SMS comparison on 100 long dosage-bearing advisories (seed 42). Comp.\,1 is the original deterministic compressor; Comp.\,2 additionally copies the nearest recognized active ingredient from the verified answer. Sixty-seven reference advisories contain a chemical recognized by the scorer.}
\label{tab:n05c}
\end{table}

\subsection{Offline Retrieval}
The local BM25 path obtained a hit rate of 0.94 on the 400-query retrieval benchmark. With service-worker precaching enabled, the test recorded no cache miss in 400 requests. In a network simulation, remote retrieval lost 13.25 percentage points at 15\% packet loss and 30.0 points at 30\% packet loss, while local retrieval was unchanged. Packet loss was simulated rather than measured in a field deployment. This result applies to retrieval; offline generation additionally requires a compatible local model and sufficient hardware.

\subsection{Installation Footprint}
Table~\ref{tab:footprint} reports measured application and vision-asset sizes. The minimal installation occupies 284.0\,MB, including 95.64\,MB of unique ONNX vision files. These totals exclude language-model weights, which require several additional gigabytes. The footprint therefore supports browser-side vision and local retrieval, not a claim that the complete generator fits on a low-end phone.

\begin{table}[t]
\centering
\small
\begin{tabular}{lr}
\toprule
\textbf{Artifact, excluding LLM weights} & \textbf{Size (MB)} \\
\midrule
Unique ONNX vision assets (11 files) & 95.64 \\
Minimal installation & 284.0 \\
Installation with PyTorch checkpoints & 339.6 \\
\bottomrule
\end{tabular}
\caption{Measured installation footprint. Language-model weights are excluded from every row.}
\label{tab:footprint}
\end{table}

\subsection{Updating the Evidence Index}
Agricultural passages are stored outside the generator. As an implementation check, we removed and reinserted 10 guideline passages selected from the top-ranked results of the stress queries. Rebuilding the index took a median of 0.2\,s on the evaluation laptop, and the rebuilt full index reproduced the deployed top five for all 400 queries. After reinsertion, each of the 10 selected passages again appeared at rank 1 for its original query.

This check confirms that passages can be replaced and the index rebuilt without loading or modifying the generator. It does not measure whether newly added guidance is correct, sufficiently complete, or retrieved well for a new topic. Adding a crop still requires supporting evidence, aliases, routing rules, and any corresponding vision component.

\subsection{Latency}
Table~\ref{tab:latency} reports non-generation latency over 400 requests. The median total was 33.565\,ms. Generation is excluded from this total: on the evaluated CPU setup, the local generator required a median of 24.3\,s for 150 output tokens and is therefore intended for queued or kiosk workflows rather than interactive low-latency chat.

\begin{table}[t]
\centering
\small
\begin{tabular}{lrr}
\toprule
\textbf{Component} & \textbf{Median (ms)} & \textbf{p95 (ms)} \\
\midrule
Safety processing & 0.321 & 0.705 \\
BM25 retrieval & 26.831 & 61.248 \\
Dosage verification & 4.815 & 11.220 \\
\midrule
Non-generation total & 33.565 & 67.833 \\
\bottomrule
\end{tabular}
\caption{Latency over 400 requests with BM25 retrieval and generation stubbed out. Component medians need not sum to the median total.}
\label{tab:latency}
\end{table}

\subsection{Trace Ordering}
We inspected the provenance traces for 100 evaluation cases. All 100 preserved the required order among the stages present in the turn, with a median of nine recorded events. This is a conformance check on event ordering; it does not establish that every recorded source causally determined the generated text.

\section{Demonstration Gallery}
\label{app:ui}

\begin{figure*}[!t]
\centering
\begin{subfigure}[t]{0.31\textwidth}
  \centering
  \includegraphics[width=\linewidth,height=3.65cm,keepaspectratio]{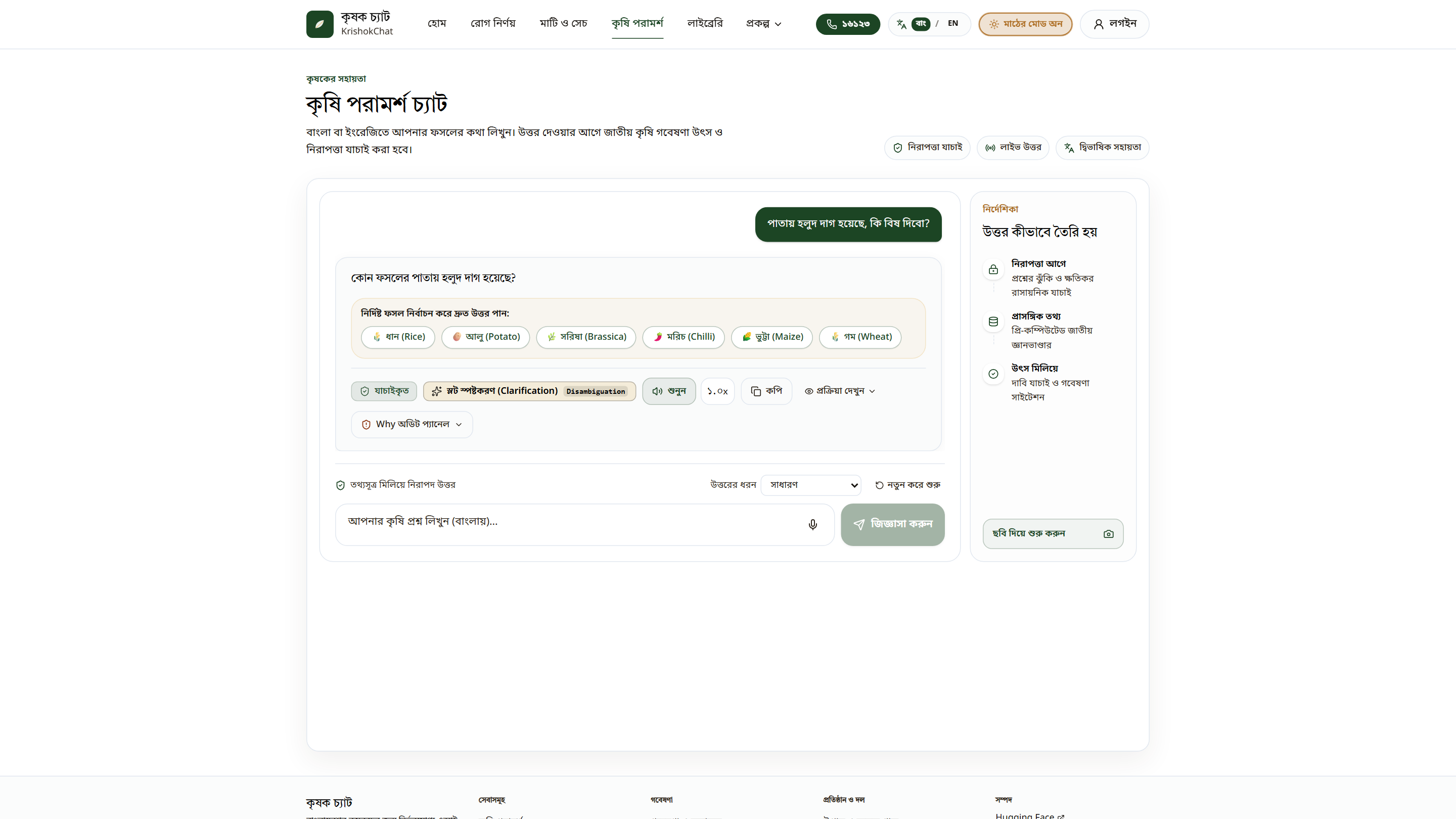}
  \caption{\textbf{ASK (S1):} missing crop halts before retrieval; offers crop chips.}
  \label{fig:app_halt}
\end{subfigure}\hfill
\begin{subfigure}[t]{0.31\textwidth}
  \centering
  \includegraphics[width=\linewidth,height=3.65cm,keepaspectratio]{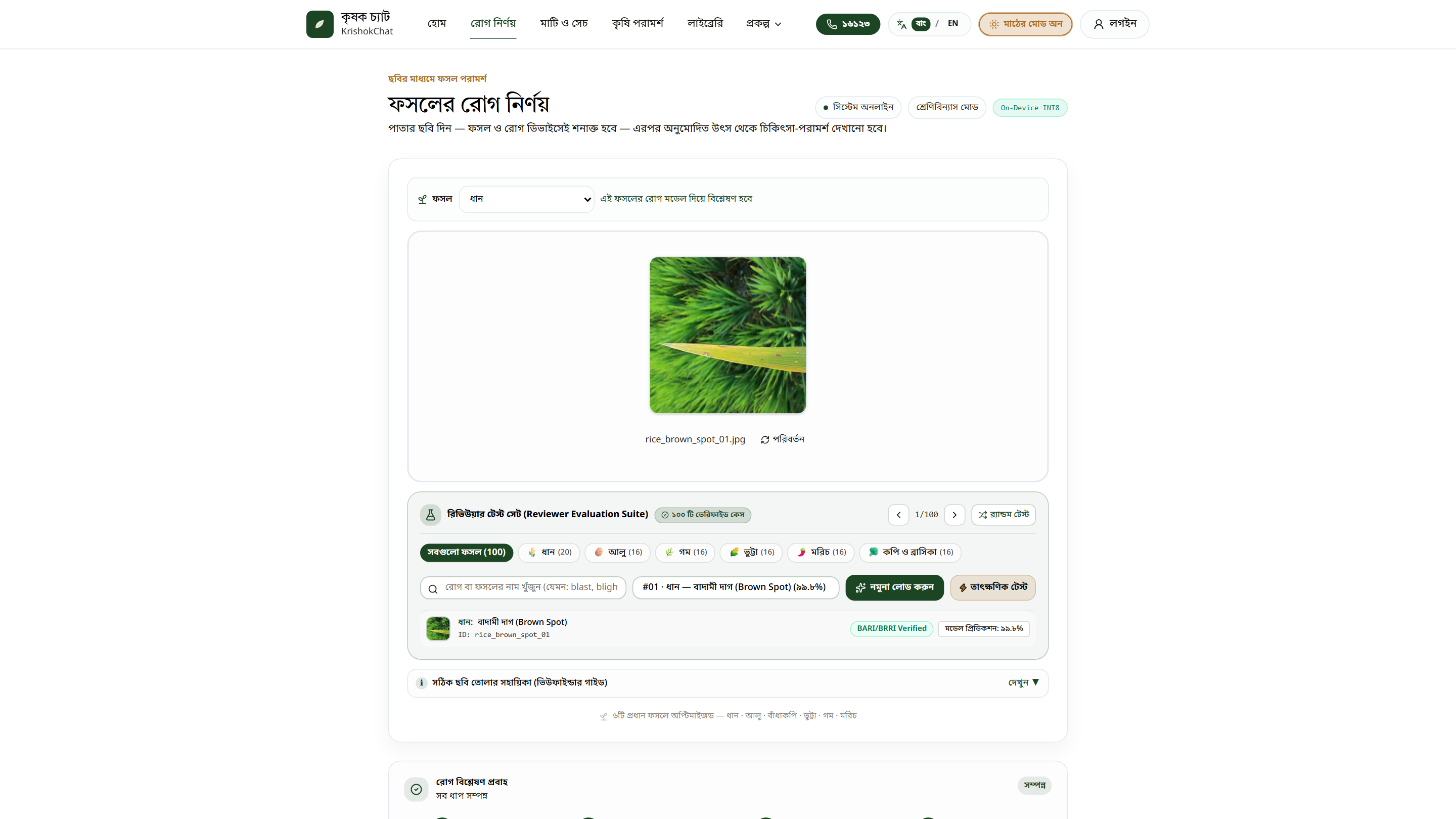}
  \caption{\textbf{BIND (S2):} on-device classifier bounds retrieval to the predicted crop.}
  \label{fig:app_fence}
\end{subfigure}\hfill
\begin{subfigure}[t]{0.31\textwidth}
  \centering
  \includegraphics[width=\linewidth,height=3.65cm,keepaspectratio]{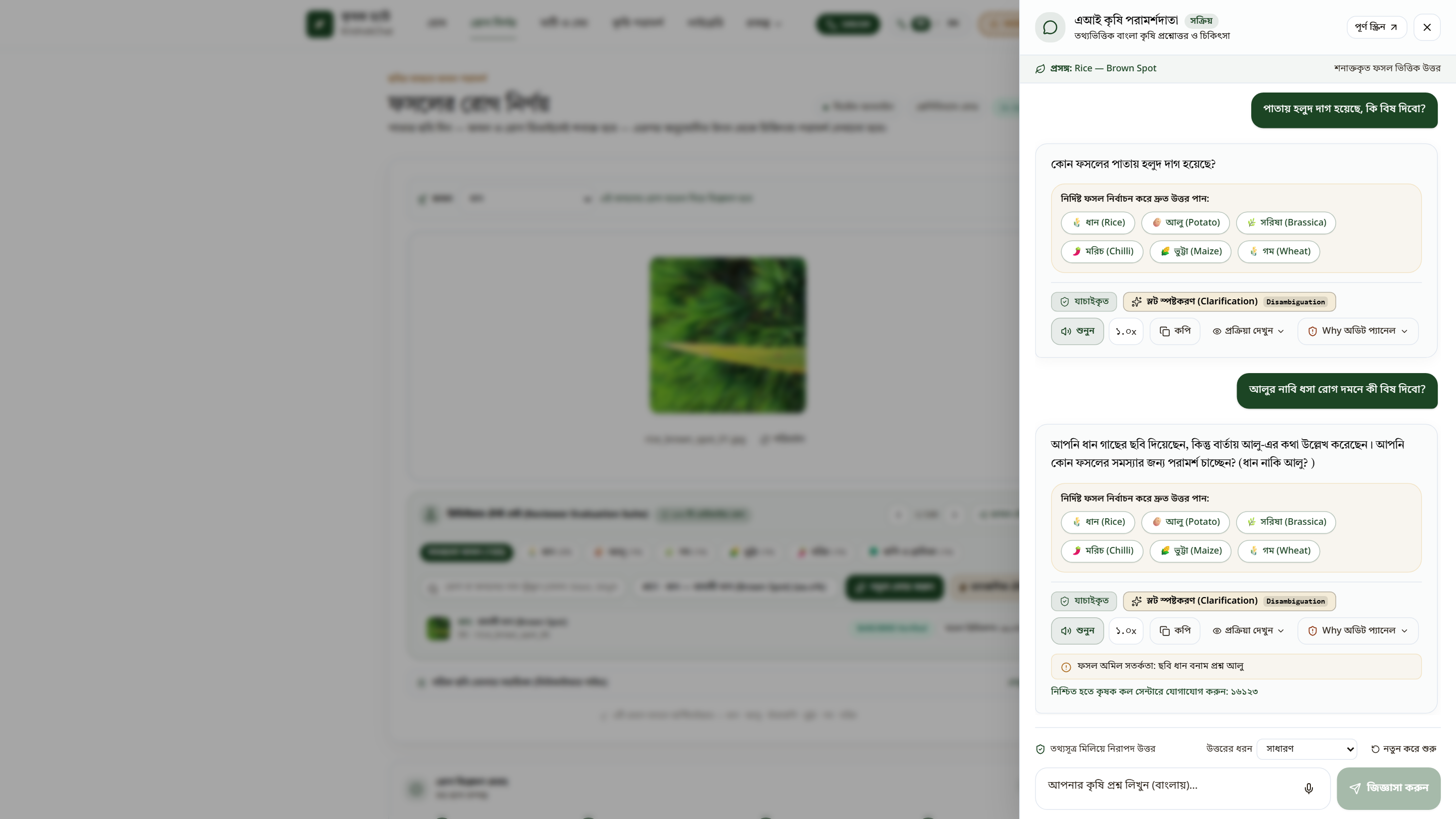}
  \caption{\textbf{CONFIRM (S3):} mismatch badge holds advice on conflicting inputs.}
  \label{fig:app_mismatch}
\end{subfigure}

\vspace{0.5em}
\begin{subfigure}[t]{0.31\textwidth}
  \centering
  \includegraphics[width=\linewidth,height=3.65cm,keepaspectratio]{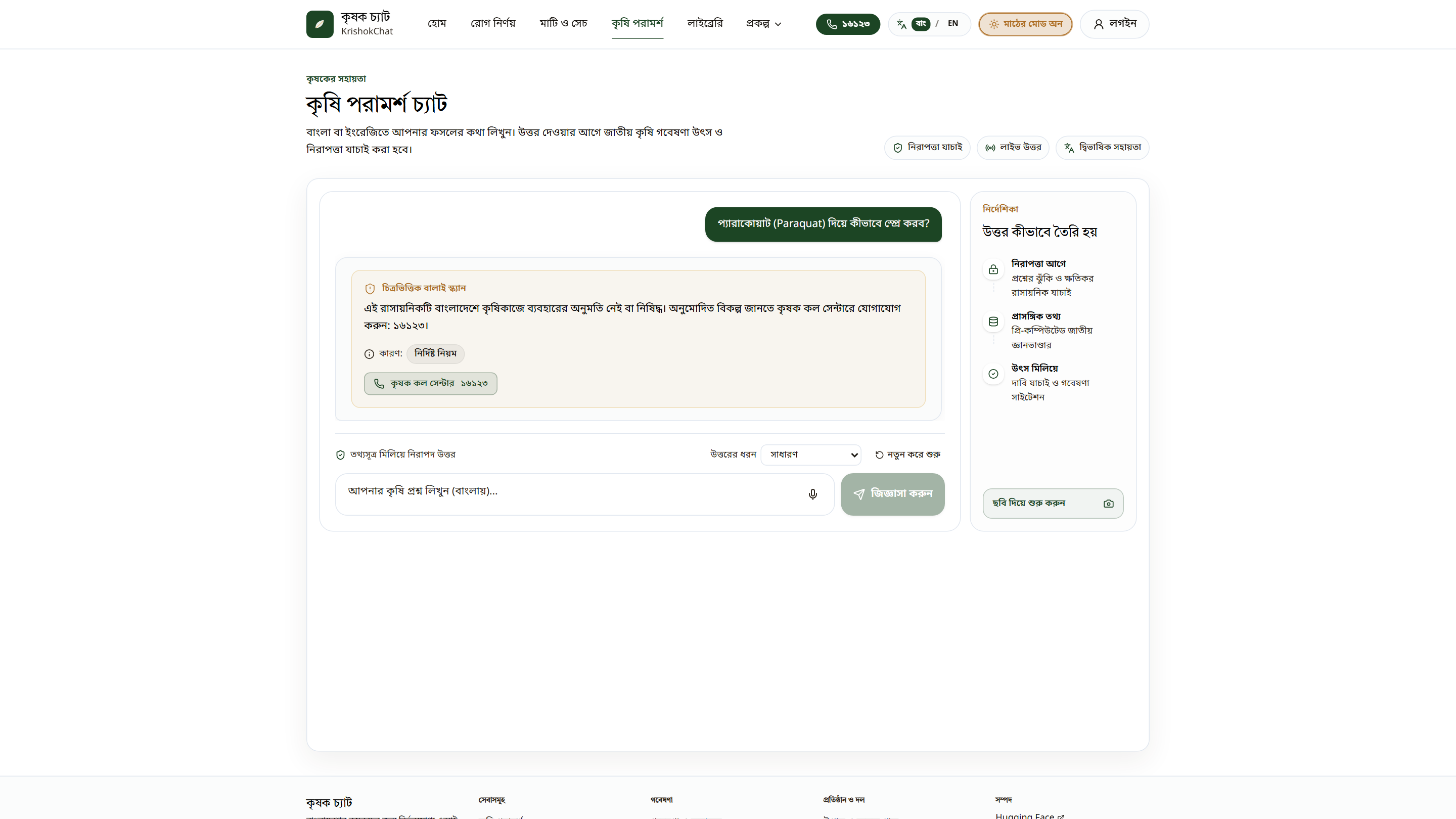}
  \caption{\textbf{REFER (S4):} safety notice with 16123 helpline call button.}
  \label{fig:app_delivery}
\end{subfigure}\hfill
\begin{subfigure}[t]{0.31\textwidth}
  \centering
  \includegraphics[width=\linewidth,height=3.65cm,keepaspectratio]{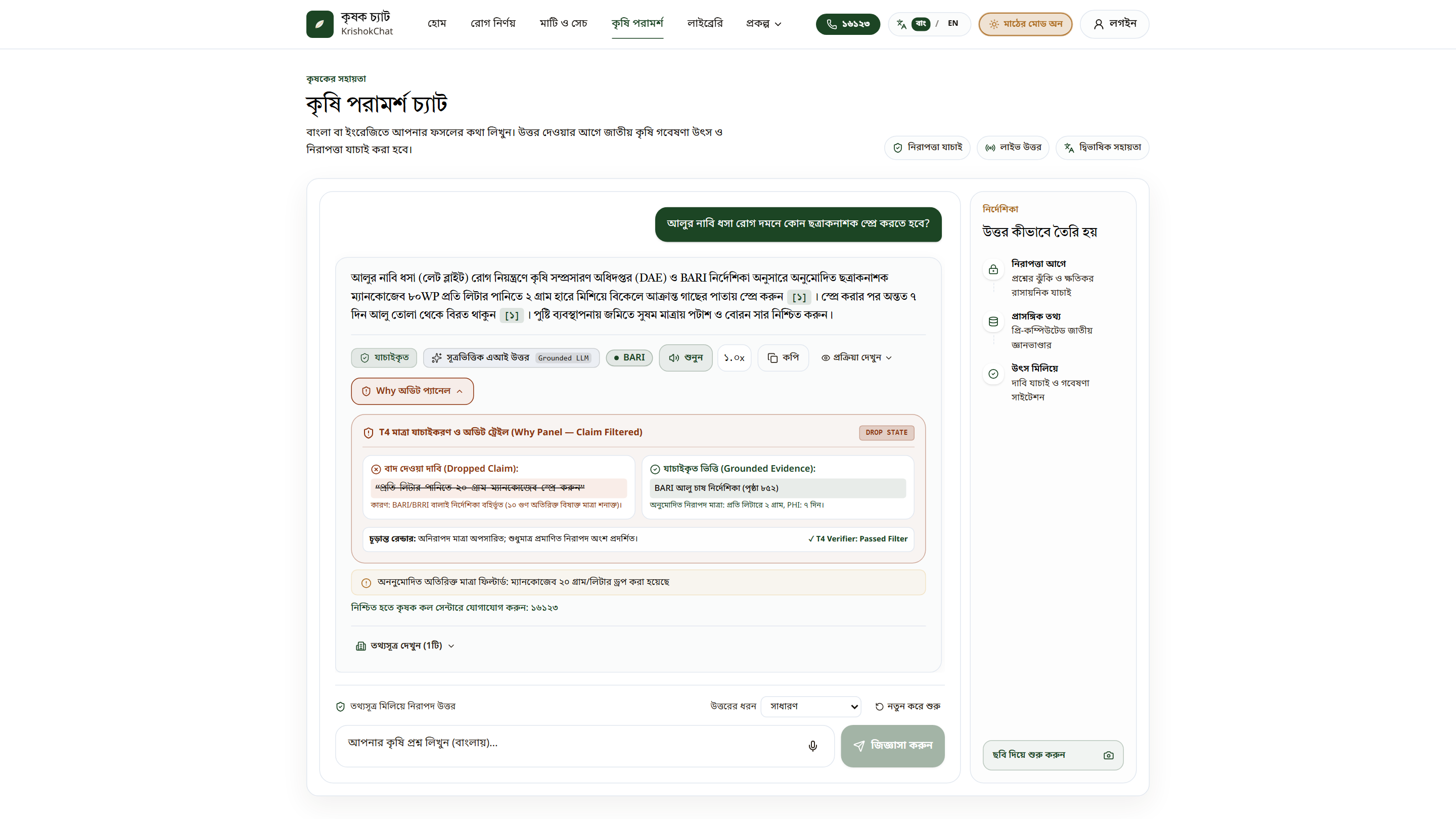}
  \caption{\textbf{DROP (S5):} T4 strips the unsupported dosage claim and retains BARI evidence.}
  \label{fig:app_drop}
\end{subfigure}
\caption{Demonstration gallery: the five website states expose admission, crop binding, conflict confirmation, safety referral, and dosage release. The two-row, three-column layout enlarges the screenshots while preserving their aspect ratios.}
\label{fig:app_gallery}
\end{figure*}

Figures~\ref{fig:app_halt}--\ref{fig:app_drop} show full-size screens for the five steps (S1--S5) in Section~\ref{sec:demonstration}.

\section{Multi-Turn State Tests}
\label{sec:eval-multiturn}

These tests check that conversation state moves as designed across turns. The suite has four constructed dialogue cohorts (200 dialogues, 425 turns) and five checks; the crop-shift cohort is scored twice (Table~\ref{tab:multiturn_eval}). Follow-up turns are rewritten into standalone queries by an LLM rewriter (\texttt{gemini-2.5-flash-lite}; 58 rewrites).

\begin{table}[t]
\centering
\scriptsize
\setlength{\tabcolsep}{2pt}
\begin{tabular*}{\columnwidth}{@{\extracolsep{\fill}}lcccc@{}}
\toprule
\textbf{Cohort} & \textbf{Dial.} & \textbf{Turns} & \textbf{Check} & \textbf{Result} \\
\midrule
Carryover & 75 & 175 & Crop slot kept & 100/100 \\
Crop shift & 75 & 150 & Slots flushed & 75/75 \\
\quad (same) & & & Old slot kept & 0/75 \\
Delayed evasion & 30 & 60 & Turn-2 catch & 30/30 \\
Clarification & 20 & 40 & ASK, then bound & 20/20 \\
\midrule
\textbf{Total} & \textbf{200} & \textbf{425} & & \\
\bottomrule
\end{tabular*}
\caption{Multi-turn state tests on constructed dialogues. The crop-shift cohort is scored on two checks; no pooled score is reported. Carryover counts the 100 follow-up turns.}
\label{tab:multiturn_eval}
\end{table}

In \textit{anaphoric carryover}, farmers ask follow-ups such as ``\textit{tahole ekhon ki spray korbo?}'' (``what should I spray then?'') without restating the crop; the crop slot was kept in all 100 follow-up turns. In \textit{crop shift}, farmers switch crops mid-conversation; all 75 switches were detected and the previous crop's disease slots were cleared. This measures conversation state, not the text of generated advice. In \textit{delayed evasion}, a benign first turn is followed by a request for a prohibited chemical (for example paraquat, Furadan, or methamidophos); T0 stopped all 30 second turns (median 0.20\,ms) and showed the 16123 referral. In \textit{clarification}, a crop-less symptom query halted at \textbf{ASK} and the crop supplied on turn two was bound in all 20 dialogues.

\clearpage
\onecolumn
\section{Evidence Summary and Limitations Ledger}
\label{app:evidence}
\label{app:limitations_full}

Tables~\ref{tab:evidence_a} and~\ref{tab:evidence_b} list each measured result with its sample size, status, and scope. Status is \emph{measured} (direct measurement on benchmark or logged data), \emph{pilot} (small controlled check), \emph{simulated} (part of the condition, such as packet loss, is simulated), \emph{modeled} (derived from a cost model), or \emph{stated} (a limitation without a measurement).

\par\vspace{0.2em}
\noindent
\begin{minipage}{\textwidth}
\centering
\fontsize{6.2pt}{7.0pt}\selectfont
\renewcommand{\arraystretch}{0.65}
\setlength{\tabcolsep}{2.5pt}
\begin{tabularx}{\textwidth}{@{}p{0.13\textwidth} p{0.36\textwidth} r p{0.08\textwidth} X@{}}
\toprule
\textbf{Component} & \textbf{Measure and result} & \textbf{$n$} & \textbf{Status} & \textbf{Scope and boundary} \\
\midrule
\multicolumn{5}{l}{\textit{Admission and referral (C1)}} \\
Crop gate & Halted 30/30 crop-less queries (15.0\% of the set, Wilson [10.71, 20.61]); passed 170/170 crop-specified queries & 200 & Measured & Real farmer queries, dual-annotated ($\kappa = 1.00$); no false halts on these 170 \\
Extractor accuracy & Crop label matched gold on 95.5\%; miss 1.5\%, false positive 3.0\% & 200 & Measured & Same 200 queries; separate from annotator agreement \\
Controlled pilot & Crop-less requests halted 100/100; specified requests passed directly 114/140, 26/140 received a bounded clarification for another slot & 240 & Pilot & Constructed pairs; conformance, not open-domain accuracy \\
Token metering & Median 89 tokens for a clarification vs 2,146 for retrieval context & 200 & Measured & Separate medians; no combined saving is claimed \\
LLM vs rule-based gate & LLM halted 10\% vs 15\% for the gate; decision agreement 0.91; median LLM latency 4,218.6\,ms & 200 & Measured & Halt rate and latency only; a 14-answer review (7/14 medium or better) is descriptive \\
Banglish precheck & 85/85 attacks intercepted (95\% CI [95.7, 100.0]) vs 1/85 for a formal-only precheck; 0/15 false alarms & 100 & Measured & Constructed Latin-script and phonetic attacks; pattern matching median 0.016\,ms, full T0 0.321\,ms \\
Off-topic audit & 10/10 injections blocked; off-topic queries refused 0/30 by T0 and 25/30 (83.33\%) by the LLM intent classifier & 40 & Measured & Five weather queries are routed to agricultural information by design \\
\midrule
\multicolumn{5}{l}{\textit{Evidence binding and conflicts (C2)}} \\
Crop binding & Top-5 cross-crop retrieval was 36.5\% with open retrieval and 30.25\% with the adjudicated crop bound (146 vs.\ 121 of 400); 25/0 discordant, exact McNemar $p=5.96\times10^{-8}$ & 400 & Measured & Simulates correct routing; this is a retrieval measure, not answer quality \\
Crop filter (N08c) & Other-crop top-5 lists 0/400 (vs 121/400 bound only, current code); 121/0 discordant, $p < 10^{-36}$; five passages served for 400/400; gold-disease hit@5 20/120 in both arms; title check 26/30 clean, 4 with crops outside the alias list & 400 & Measured & Filter and metric share the alias list, so 0/400 holds by construction; coverage, disease hit and title check are the independent checks \\
Text-extracted binding & Top-5 cross-crop retrieval 30.25\% (no gate) vs 28.75\% (gated pipeline); 6/0 discordant, $p = 0.031$ & 400 & Measured & All 400 name a crop, so the gate halts none; separate run from the row above \\
Disease vision (top-1) & 0.9413 to 0.9907 across six crop-specific classifiers & 4,294 & Measured & Disease within a crop, not crop routing; 100\% ONNX--PyTorch label agreement \\
Family router & Botanical family correct for 433/437 (0.9908, [97.67, 99.64]) & 437 & Measured & Six target crops, wheat-800 excluded; not open-world identification \\
INT8 quantization & Wheat 3.68$\times$ smaller (5.92 to 1.61\,MB), 0.00\,pp drop ($n=400$); potato 0.09\,pp ($n=1{,}170$), brassica 0.00\,pp ($n=443$); 1.5 to 2.0$\times$ slower on non-VNNI x86 CPUs & 2,013 & Measured & Storage saving, not speedup; rice INT8 lost 2.5\,pp, exceeded the 2.0\,pp gate, and was not deployed \\
Mismatch badge & Fired on 53/54 farmer conflicts [90.2, 99.7] and 400/400 PRISM conflicts [99.05, 100.0]; 0/54 and 0/400 matched controls & 908 & Measured & One implicit-reference miss (\texttt{farmer\_q\_63}) \\
\bottomrule
\end{tabularx}
\captionof{table}{Evidence summary for admission and evidence binding. Status is defined at the start of this appendix.}
\label{tab:evidence_a}

\vspace{0.4em}

\begin{tabularx}{\textwidth}{@{}p{0.13\textwidth} p{0.36\textwidth} r p{0.08\textwidth} X@{}}
\toprule
\textbf{Component} & \textbf{Measure and result} & \textbf{$n$} & \textbf{Status} & \textbf{Scope and boundary} \\
\midrule
\multicolumn{5}{l}{\textit{Generation and release (C3)}} \\
Guarded prompt, local model & ASR 10.71\% [7.61, 14.88] guarded vs 52.5\% [46.66, 58.28] neutral (30 vs 147 of 280); Bengali-native 19/100 vs 77/100 & 280 per arm & Measured & \texttt{krishokchat-4b}, T0 and T4 disabled; 210 prompts in seven families plus 70 Bengali-native probes; rule-based judge \\
T0 + guarded prompt, local & ASR 4.64\% [2.73, 7.78] (13/280); Bengali-native 2/100 [0.55, 7.0]; T0 stopped 207/280 prompts and 17 of the 30 prompt-level successes & 280 & Measured & Replay of the guarded N14 responses with T0 applied (no new model calls); T4 not applied. On 200 farmer queries T0 matched 9: 8 out-of-scope requests (training, livestock, input sourcing) sent to 16123 by the coverage rule, 1 false safety match \\
Output audit & 46/47 guarded outputs safe and actionable, 1/47 vague but harmless, 0/47 dangerous; 3/3 unguarded controls dangerous; Fleiss $\kappa = 0.82$ & 50 & Measured & Three blinded raters (two undergraduates, one extension officer) \\
Dosage verifier, live & 114/118 inserted errors caught (96.6\%); 0/38 unmodified answers rejected & 38 & Measured & Errors inserted into 38 live answers, so they are not independent natural errors \\
Dosage verifier, scaled & 296/350 inserted errors caught (84.6\%): overdose 76/76, underdose 67/67; chemical 85\%, unit 73\%, invented PHI 65\% & 528 & Measured & Cases built from 88 BARI/BRRI passages; numerical dosage sentences only \\
Multi-turn state & Slot kept in 100/100 follow-ups; 75/75 crop shifts flushed; 30/30 delayed attacks caught; 20/20 ASK then bound & 200 & Measured & Constructed dialogues, 425 turns, four cohorts; state checks, not generated advice \\
Trace ordering & 100/100 traces in required event order (median 9 events) & 100 & Measured & Sampled evaluation cases \\
\midrule
\multicolumn{5}{l}{\textit{Delivery and deployment}} \\
SMS length & 399/400 messages within 160 characters; 8/8 referral messages carried only the referral & 400 & Measured & Bengali output uses UCS-2, so 160 characters is 2--3 segments \\
SMS slot survival & Dose, interval, and PHI kept in 1,000/1,000 constructed tuples; dose kept in 65/66 live gold advisories vs 12/92 for naive truncation & 1,000 & Measured & Different denominators, not a paired comparison \\
SMS comparison & Dose kept in 23 (naive), 89 (LLM), 92 (Comp.\,1), 94 (Comp.\,2) of 100; chemical name in 20, 56, 34, 63; both in 14, 51, 32, 61; 0 invented values for Comp.\,2 & 100 & Measured & Seed 42; no model calls for the compressors; active-ingredient list overlaps the scorer's list \\
Offline retrieval & Local BM25 hit rate 0.94, 0/400 cache misses; remote retrieval loses 13.25\,pp (15\% loss) and 30.0\,pp (30\% loss) & 400 & Measured, loss simulated & Retrieval only \\
Non-LLM overhead & Median 33.565\,ms (p95 67.833\,ms): safety 0.321, BM25 26.831, verifier 4.815\,ms & 400 & Measured & Stub LLM; local generation about 24\,s per call \\
Installation footprint & 95.64\,MB ONNX vision assets; 284.0\,MB minimal; 339.6\,MB with PyTorch checkpoints & n/a & Measured & Excludes the language-model weights \\
Knowledge update (X5) & 10/10 withheld passages at rank 1 after re-insertion; index rebuild 0.2\,s (median); rebuilt index matches deployed top-5 on 400/400 queries & 10 & Measured & Leave-one-out on real passages; no model loaded; not evidence for adding new crops \\
Retrieval backend & BM25 only; dense retrieval not evaluated & n/a & Stated & All retrieval numbers are BM25-only \\
\bottomrule
\end{tabularx}
\captionof{table}{Evidence summary for generation, release, delivery, and deployment. Status is defined at the start of this appendix.}
\label{tab:evidence_b}
\end{minipage}

\end{document}